\documentclass[conference]{IEEEtran}
\IEEEoverridecommandlockouts

\usepackage{cite}
\usepackage{amsmath,amssymb,amsfonts}
\usepackage{graphicx}
\usepackage{textcomp}
\usepackage{xcolor}

\usepackage{booktabs}
\usepackage{multirow}
\usepackage{diagbox}
\usepackage[colorlinks,linkcolor=blue]{hyperref}

\usepackage{booktabs}
\usepackage{soul}
\usepackage{makecell}
\usepackage{bbding}
\usepackage{pifont}
\usepackage{bigstrut}
\usepackage{color}
\usepackage{algorithm}
\usepackage{algpseudocode}
\usepackage{multirow}

\def\BibTeX{{\rm B\kern-.05em{\sc i\kern-.025em b}\kern-.08em
    T\kern-.1667em\lower.7ex\hbox{E}\kern-.125emX}}
\begin{document}



\title{GPF-Net: Gated Progressive Fusion for Polyp Re-Identification
\thanks{* Corresponding author. This work was partially supported by the National Natural Science Foundation of China under Grant No.62301315 and 2025 Key Research Initiatives of Yunnan Erhai Lake National Ecosystem Field Observation Station under Grant No.2025ZD03.}
}

\author{\IEEEauthorblockN{1\textsuperscript{st} Suncheng Xiang*}
\IEEEauthorblockA{\textit{Shanghai Jiao Tong University} \\
Shanghai, China \\
xiangsuncheng17@sjtu.edu.cn}
\and
\IEEEauthorblockN{2\textsuperscript{nd} Xiaoyang Wang}
\IEEEauthorblockA{\textit{Peking University} \\
Beijing, China \\
xiaoyangwang25@stu.pku.edu.cn}
\and
\IEEEauthorblockN{3\textsuperscript{rd} Junjie Jiang}
\IEEEauthorblockA{\textit{Shanghai Fifth People’s Hospital} \\
Shanghai, China \\
xnjh121644@163.com}
\and
\IEEEauthorblockN{4\textsuperscript{th} Hejia Wang}
\IEEEauthorblockA{\textit{Shanghai Jiao Tong University} \\
Shanghai, China \\
he-jacklia-jia@sjtu.edu.cn}
\and
\IEEEauthorblockN{5\textsuperscript{th} Dahong Qian}
\IEEEauthorblockA{\textit{Shanghai Jiao Tong University} \\
Shanghai, China \\
dahong.qian@sjtu.edu.cn}
}

\maketitle

\begin{abstract}
Colonoscopic Polyp Re-Identification (ReID) aims to match the same polyp across a large gallery of images captured from different viewpoints and with different cameras, playing a critical role in computer-aided diagnosis for the prevention and treatment of colorectal cancer. However, the coarse granularity of high-level features often limits performance on small polyps, where fine-grained details are essential for accurate matching. To address this challenge, we propose a novel multimodal feature fusion architecture, termed the Gated Progressive Fusion Network, which selectively integrates features from multiple levels through fully connected gating mechanisms. Building on this framework, we introduce a gated progressive fusion strategy that enables layer-wise refinement of semantic information, facilitating multi-level feature interactions to enhance both generalization ability and robustness. Extensive experiments on standard benchmarks demonstrate the advantages of the multimodal setting over state-of-the-art unimodal ReID models, particularly when combined with the proposed fusion strategy tailored for general-purpose scenarios.
\end{abstract}

\begin{IEEEkeywords}
Colonoscopic polyp re-Identification, modal representation, gated progressive fusion.
\end{IEEEkeywords}

\section{Introduction}
\label{sec1}

\begin{figure}[!t]
    \centering
    \includegraphics[width=0.5\textwidth]{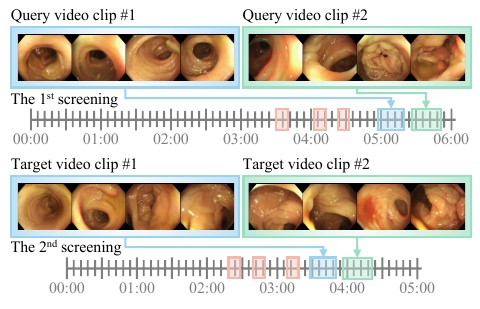}
    \caption{We introduce a multimodal polyp dataset for polyp ReID, where each polyp ReID will encounter many challenges such as variation in terms of backgrounds, viewpoint, and illumination.}
    \label{fig1}
\end{figure}

Colonoscopic polyp re-identification (Polyp ReID) aims to match a specific polyp in a large gallery with different cameras and locations, which has been studied intensively due to its practical importance in the prevention and treatment of colorectal cancer in the computer-aided diagnosis. With the development of deep convolution neural networks and the availability of video re-identification dataset, video retrieval methods have achieved remarkable performance in a supervised manner~\cite{feng2019spatio}, where a model is trained and tested on different splits of the same dataset. However, in practice, manually labeling a large diversity of pairwise polyp area data is time-consuming and labor-intensive when directly deploying polyp ReID system to new hospital scenarios~\cite{chen2023colo,xiang2024vt}. Nevertheless, when compared with the conventional ReID, polyp ReID is confronted with more challenges in some aspects: \textbf{1) from the model perspective:} traditional object ReID methods learn the unimodal representation by greedily “pre-training” several layers of features on the basis of visual samples, while ignore to explore complementary information from different modalities, and \textbf{2) from the data perspective}, polyp ReID  will encounter many challenges such as variation in terms of backgrounds, viewpoint, and illumination, \textit{etc.}, as shown Figure~\ref{fig1}, which poses great challenges to the clinical deployment of deep model in real-world scenarios.

Our key innovation lies in the Gated Progressive Fusion Network (GPF-Net), which combines a gated attention mechanism inspired by Gated Multimodal Units with a progressive fusion strategy, which iteratively refines multimodal representations through four stacked gated fusion layers. In contrast to traditional single-stage fusion, which often forces a monolithic integration of heterogeneous data, our approach introduces a novel coarse-to-fine alignment mechanism. It begins by disentangling low-level discriminative features, such as texture and spectral signatures, and then hierarchically bridges them with high-level semantic constructs, including object categories and relational contexts. This progressive integration not only preserves fine-grained details but also mitigates cross-modal interference, directly rectifying the inherent limitations of existing end-to-end fusion paradigms. Extensive experiments on polypus dataset demonstrate that GPF-Net achieves state-of-the-art performance, outperforming existing methods by significant margins in mAP and Rank-1 accuracy. Our work not only advances the field of polyp ReID but also provides insights into effective multimodal fusion strategies for medical image retrieval tasks.

In summary, the main contributions of this paper can be summarized as follows: 1) We construct a multimodal feature fusion framework named GPF-Net majoring in multi-scale feature extraction for polyp ReID. 2) A dynamic gating progressive fusion mechanism is introduced to achieve layer-wise refinement of semantic information through multi-level feature interactions, which can adaptively adjust modality weights based on visual content, then enhancing feature discriminability. 3) Experiments demonstrate that the proposed method significantly outperforms existing approaches across multiple metrics, while maintaining competitive computational complexity than current polyp ReID methods.

\section{Related Work}
\label{sec2}
\subsection{Computer-Aided Colonoscopy Analysis}

Recent advances in deep learning have significantly improved computer-aided diagnosis (CAD) systems for colonoscopy~\cite{zhou2025ldp}. Existing studies mainly focus on polyp detection and segmentation tasks. For example, Pacal \textit{et al.}~\cite{pacal2022yolo} developed a real-time polyp detection framework based on YOLO, while Puyal \textit{et al.}~\cite{puyal2022hybrid} exploited temporal information through a hybrid 2D/3D convolutional network for video-based polyp detection. In the segmentation domain, attention-based architectures~\cite{wei2021sanet} and multi-scale feature learning strategies~\cite{zhao2021msnet} have demonstrated strong performance in accurately delineating polyp boundaries. Although these methods substantially improve lesion discovery and characterization, they primarily address single-examination scenarios and do not explicitly consider the challenging task of matching the same polyp across multiple colonoscopy procedures.

\subsection{Polyp Re-Identification}

Polyp re-identification (Polyp ReID) aims to retrieve the same lesion from different colonoscopic examinations of the same patient. Compared with conventional retrieval tasks, Polyp ReID is complicated by large appearance variations caused by viewpoint changes, illumination differences, bowel deformation, and tissue evolution. To facilitate research in this area, Chen \textit{et al.}~\cite{chen2023colo} introduced a large-scale benchmark dataset and proposed Colo-SCRL, which leverages self-supervised contrastive learning for colonoscopic video retrieval. Subsequently, Xiang \textit{et al.}~\cite{xiang2025meta} explored meta-learning strategies to improve representation discriminability and generalization capability. These studies established the foundation of Polyp ReID and demonstrated the effectiveness of representation learning for polyp retrieval.

\subsection{Vision-Language Multimodal Learning}

Recent progress in multimodal learning has shown that textual descriptions can provide complementary semantic information beyond visual appearance. The success of CLIP~\cite{radford2021clip} has inspired numerous medical vision-language models, such as MedCLIP~\cite{wang2022medclip}, BioViL~\cite{boecking2022biovil}, and PubMedCLIP~\cite{ewuoso2023pubmedclip}. Motivated by the effectiveness of multimodal representation learning, recent studies have begun incorporating textual clinical information into Polyp ReID systems. VT-ReID~\cite{xiang2024vt} first incorporated colonoscopy reports into the retrieval framework and employed contrastive learning to enhance visual-text alignment. More recently, DMCL~\cite{xiang2024learning} proposed a deep multimodal collaborative learning framework with dynamic feature fusion mechanisms to jointly optimize visual and textual representations.




Despite these achievements, existing multimodal Polyp ReID methods still face two major limitations. First, most approaches rely on shallow fusion strategies, which may not fully exploit the complementary semantic information across heterogeneous modalities. Second, increasingly complex architectures often introduce substantial computational overhead and parameter redundancy, limiting their deployment in practical clinical environments. To address these issues, we propose a Gated Progressive Fusion Network (GPF-Net), which employs dynamic modality weighting and progressive cross-modal interaction to achieve more effective and parameter-efficient visual-text fusion for polyp re-identification.

\section{Methodology}
\label{sec3}

\subsection{Preliminary}
We begin with a formal description of the colonoscopic polyp re-identification (Polyp ReID) problem. Assuming that we are given a source domain $\mathcal{D}$, which contains its own image-label pairs $\mathcal{D}=\left\{\left(\boldsymbol{x}_i, y_i\right)\right\}_{i=1}^{N}$ of colonoscopic videos, where  $N$ is the number of images in the source domain $ \mathcal{D}$. Each sample $\boldsymbol{x}_i \in \mathcal{X}$ is associated with an identity label $y_i \in \mathcal{Y}=\left\{1,2, \ldots, M\right\}$, where $M$ is the number of identities in the source domain $\mathcal{D}$. The goal of this work is to leverage labeled source training polyp samples to learn the discriminative embeddings of the target testing set on polyp ReID task.


\begin{figure*}[!t]
    \centering
    \includegraphics[width=0.95\textwidth]{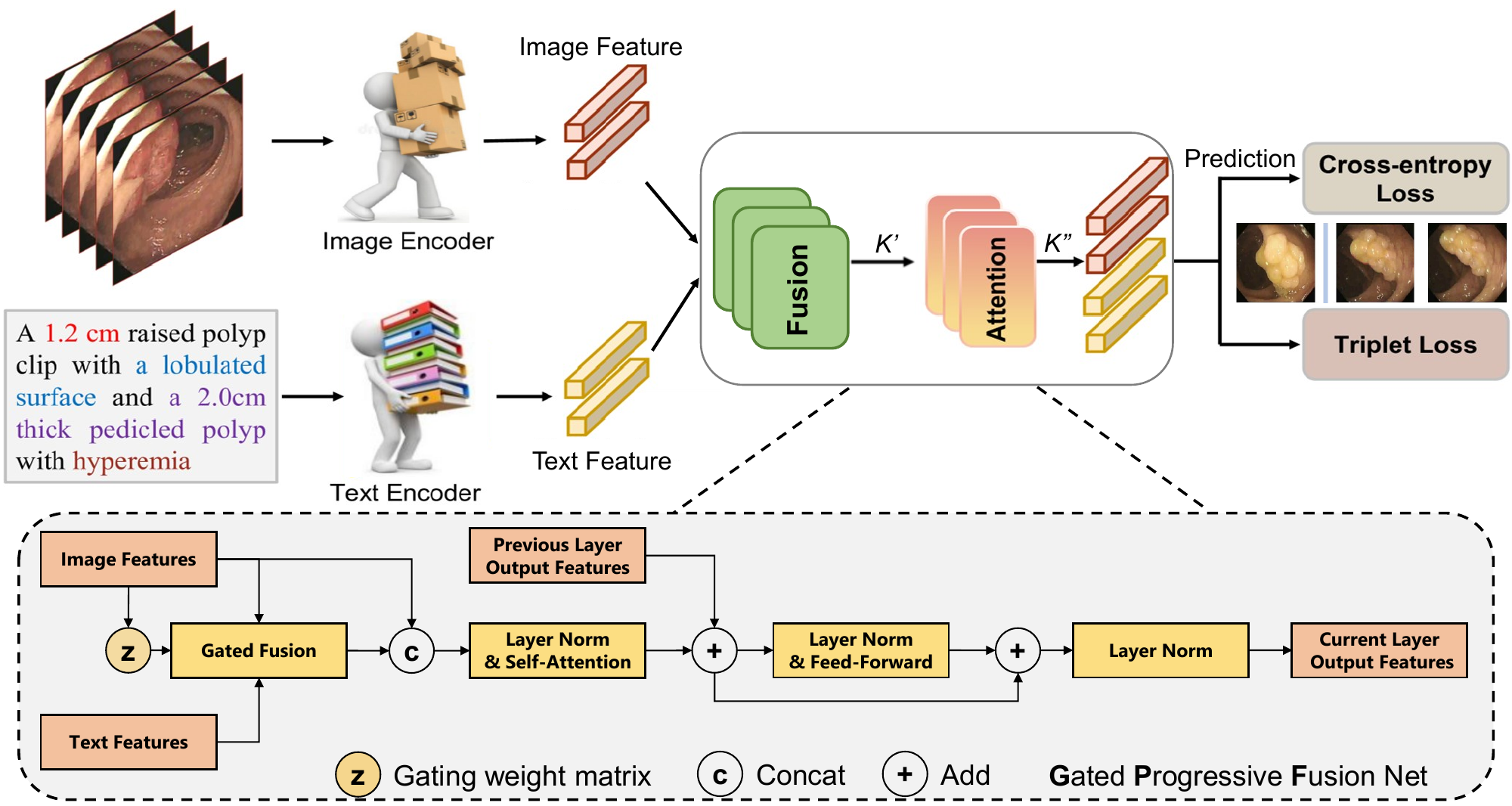}
    \caption{The overall architecture of our GPF-Net is designed around two core components: a visual-textual feature extraction model and a two-stage feature fusion module, which work in tandem to process multi-modal data.}
    \label{fig2}
\end{figure*}

\subsection{Our Proposed GPF-Net Framework}

The overall architecture of our model is illustrated in Figure~\ref{fig2}, which primarily consists of: (1) a visual-textual feature extraction module, and (2) a two-stage feature fusion module.
Specially, the model in this study employs a pre-trained ResNet-50 as the image feature extraction network. Specifically, given a polyp image with dimensions 224×224×3, after normalization, the image feature extraction network encodes the input into a 1×2048-dimensional feature vector. This vector is then dimensionally reduced to obtain a 512-dimensional image feature $I \in \mathbb{R}^{d_v}$. For text feature extraction, the study utilizes a pre-trained ALBERT model~\cite{albert2019lite} as the text feature extraction network. Given a colonoscopy report description containing $n$ tokens, ALBERT encodes it into an n $\times$ 768-dimensional text feature $T \in \mathbb{R}^{d_t}$.

To effectively integrate visual and textual features, we propose a Gated Progressive Fusion Network combined with a dual-stage fusion strategy based on a Transformer encoder. Specifically, the image feature $I$ and the text feature $T$ are concatenated, and positional encoding is applied to this concatenated result to obtain the preliminary fused feature $[I, T] \in \mathbb{R}^{2 d}$. Subsequently, the extracted feature $[I, T]$  is fed into a 4-layer gated progressive fusion network. Through the Gating Progressive Fusion mechanism, the contribution weights of visual and textual features are dynamically adjusted. Additionally, the progressive fusion approach avoids a single-step direct fusion of the two modalities, ultimately generating the intermediate feature $K^{\prime}$ through multiple rounds of interactive fusion between the visual and textual feature.

Finally, the fused feature $K^{\prime \prime}$ output by Transformer is employed to compute the model loss for optimizing the model $\theta$. Regarding the design of the specific loss function, this work considers that many previous studies~\cite{xiang2023rethinking} have demonstrated the importance of employing multiple loss functions for training robust and generalizable ReID models. Therefore, during training we adopt identity loss and triplet loss as the optimization objectives of our model.
Specifically, the triplet loss function aims to reduce the feature distance between similar polyp images while increasing the feature distance between dissimilar ones. Meanwhile, the identity loss function  enhances the model's ability to recognize and classify polyp categories. 


\subsection{Dynamic Gating Progressive Fusion Mechanism}

On the basis of GPF-Net, we also introduce the gating progressive fusion mechanism to achieve layer-wise
refinement of semantic information through multi-level feature interactions. First, the image features $I$, obtained through the visual feature extraction network ResNet-50~\cite{he2016deep}, are passed through a fully connected layer to produce an intermediate vector $W_z \cdot I$, where $W_z$  is a learnable matrix. Subsequently, this intermediate vector $W_z \cdot I$ is processed through a Sigmoid activation function, thereby generating the gating weight matrix $z$, shown in Eq.~\ref{eq1}. Finally, the generated gating weight matrix $z$ is utilized to adjust the proportions of the image and text modalities in the fused features, as following:
\begin{equation}
z \leftarrow \operatorname{Sigmoid}\left(W_z^i \cdot I\right)
\label{eq1}
\end{equation}
\begin{equation}
\label{eq2}
K^{\prime} \leftarrow z * T+(1-z) * I
\end{equation}
\begin{equation}
\label{eq3}
K^{\prime \prime}=\operatorname{LayerNorm}\left(\left[I, K^{\prime}\right]\right)
\end{equation}
where $K^{\prime}$ represents the features after gated weighted fusion.
Furthermore, in contrast to traditional gated fusion methods that directly adopt the weighted vector as the fused feature, we leverage the characteristics of the Multi-Head Attention layer to map the fused features into a higher-dimensional feature space, which enables the thorough mining and extraction of deep semantic information from the fused vectors, resulting in the final fused feature $K^{\prime \prime}$,  as shown in Eq.~\ref{eq3}. To strike a balance between fully extracting features for accurate polyp ReID and reducing model parameters and computational load to maintain model lightweightness, we also employ an 8-head self-attention mechanism in the 4-layer gated progressive fusion network, while utilizing a 4-head self-attention mechanism in the final 4-layer Transformer encoder.

\section{Experiments}
\label{sec4}

\subsection{Datasets and Evaluation Metric}
\label{sec4.1}

We conduct experiments on several large-scale public datasets, which include Colo-Pair~\cite{chen2023colo}, Market-1501~\cite{zheng2015scalable}, DukeMTMC-reID~\cite{zheng2017unlabeled} and CUHK03 dataset~\cite{li2014deepreid}.
We follow the standard evaluation protocol~\cite{zheng2015scalable} used in the ReID task and adopt mean Average Precision (mAP) and Cumulative Matching Characteristics (CMC) at Rank-1, Rank-5, and Rank-10 for performance evaluation on downstream ReID task.

\subsection{Implementation details}
\label{sec3.2}
 Following the training procedure in~\cite{xiang2024vt}, we adopt the common methods such as random flipping and random cropping for data augmentation and employ the Adam optimizer with a weight decay co-efficient of $1 \times 10^{-5}$ and $1 \times 10^{-7}$ for parameter optimization. Besides, we adopt the ID loss and triplet loss functions to train the model for 180 iterations, and the cosine distance is also adopted to calculate the similarity of polyp features in the dataset for the task of polyp ReID. In addition, the batch size $M_{batch}$ for training is set to 64. All the experiments are performed on the PyTorch framework with one Nvidia GeForce RTX 4090 GPU on a server equipped with an AMD EPYC 7713 64-Core Processor.

\subsection{Comparison with State-of-the-Arts}

\textbf{Colonoscopic Polyp Re-Identification.}
In this section, we compare our proposed method with the state-of-the-art algorithms,
including: (1) transformer based  (soft attention) models ViT~\cite{caron2021emerging}, Colo-SCRL~\cite{chen2023colo} and VT-ReID~\cite{xiang2024vt}; (2) knowledge distillation-based methods, such as CgS$^c$, FgAttS$^f_A$ and FgBinS$^f_B$~\cite{kordopatis2022dns}; (3) feature level based methods, such as ViSiL~\cite{kordopatis2019visil}, CoCLR~\cite{han2020self}, TCA~\cite{shao2021temporal}, CVRL~\cite{qian2021spatiotemporal}. According to the results in Table~\ref{tab2}, we can easily observe that our method shows the clear performance superiority over other state-of-the-arts with significant Rank-1 and mAP advantages. For instance, when compared to the knowledge distillation-based network FgAttS$^f_A$~\cite{kordopatis2022dns}, our model improves Rank-1 accuracy by \textbf{+70.5\%} (80.2 vs. 9.7). Besides, GPF-Net also surpasses recent transformer based  (soft attention) models ViT~\cite{caron2021emerging}, Colo-SCRL~\cite{chen2023colo} and VT-ReID~\cite{xiang2024vt}. Specially, our method outperforms the second best model DMCL~\cite{xiang2024learning} by \textbf{+22.5\%} (68.9 vs. 46.4) and \textbf{+25.9\%} (80.2 vs. 54.3) in terms of mAP and Rank-1 accuracy, respectively. The superiority of our proposed method is twofold. First, it leverages the richly informative visual-textual representations learned by GPF-Net through a multi-round collaborative training paradigm. Second, it employs a dynamic multimodal feature fusion strategy that adaptively weighs cross-modal contributions. Together, these components exert a synergistic effect that yields a more robust and discriminative model for polyp ReID tasks.

\begin{table}[!t]
\centering
\caption{Performance comparison with state-of-the-art methods on Colo-Pair dataset. \textbf{Bold} indicates the best and \underline{underline} the second best.}
\footnotesize
\setlength{\tabcolsep}{1.95mm}{
\begin{tabular}{lccccc}
\toprule
\multirow{2}{*}{Method} & \multirow{2}{*}{Venue} & \multicolumn{4}{c}{Video Retrieval $\uparrow$} \\
\cmidrule{3-6}  &  & mAP & Rank-1 & Rank-5 & Rank-10  \\
\midrule
ViSiL~\cite{kordopatis2019visil} & ICCV  & 24.9 & 14.5 & 30.6 & 51.6  \\
CoCLR~\cite{han2020self} & NIPS  & 16.3 & 6.5 & 22.6 & 33.9  \\
TCA~\cite{shao2021temporal} & WCAV  & 27.8 & 16.1 & 35.5 & 53.2  \\
ViT~\cite{caron2021emerging} & CVPR  & 20.4 & 9.7 & 30.6 & 43.5  \\
CVRL~\cite{qian2021spatiotemporal} & CVPR  & 23.6  & 11.3 & 32.3 & 53.2  \\
CgS$^c$~\cite{kordopatis2022dns} & IJCV & 21.4  & 8.1 & 35.5 & 45.2  \\
FgAttS$^f_A$~\cite{kordopatis2022dns} & IJCV  & 23.6  & 9.7 & 40.3 & 50.0  \\
FgBinS$^f_B$~\cite{kordopatis2022dns} & IJCV  & 21.2  & 9.7 & 32.3 & 48.4  \\
Colo-SCRL~\cite{chen2023colo} & ICME  & 31.5  & 22.6 & 41.9 & 58.1  \\
VT-ReID~\cite{xiang2024vt} & ICASSP  & 37.9  & 23.4 & 44.5 & 60.1  \\
DMCL~\cite{xiang2024learning} & N/A  & \underline{46.4} & \underline{54.3} & \underline{57.9} & \underline{60.4} \\
\midrule
\textbf{GPF-Net} & \textbf{Ours} & \textbf{68.9} & \textbf{80.2} & \textbf{89.6} & \textbf{91.6} \\
\bottomrule
\end{tabular}}
\label{tab2}
\end{table}

\begin{table}[!t]
  \centering
  \caption{Performance comparison with other methods on Person ReID datasets. \textbf{Bold} indicates the best and \underline{underline} the second best.}
  \footnotesize
  \setlength{\tabcolsep}{1.75mm}{
    \begin{tabular}{lcccccc}
    \toprule
    \multirow{2}[4]{*}{Method} & \multicolumn{2}{c}{Market-1501} & \multicolumn{2}{c}{DukeMTMC-reID} & \multicolumn{2}{c}{CUHK03} \\
\cmidrule{2-7}          & mAP   & Rank-1 & mAP   & Rank-1 & mAP   & Rank-1 \\
    \midrule
    PCB~\cite{wang2020surpassing} & 81.6  & 93.8  & 69.2  & 83.3  & 57.5  & 63.7 \\
    MHN~\cite{chen2019mixed}   & 85.0  & 95.1  & 77.2 & 77.3  & 76.5  & 71.7 \\
    ISP~\cite{zhu2020identity}   & 88.6  & 95.3  & 80.0  & 89.6  & 71.4  & 75.2 \\
    CBDB~\cite{tan2021incomplete} & 85.0  & 94.4  & 74.3  & 87.7  & 72.8  & 75.4 \\
    C2F~\cite{zhang2021coarse}   & 87.7  & 94.8  & 74.9  & 87.4  & 84.1  & 81.3 \\
    NFormer~\cite{wang2022nformer} & 91.1  & 94.7  & \underline{83.5}  & 89.4  & 74.7  & 77.3  \\
    MGN~\cite{wang2018learning}   & 86.9  & 95.7  & 78.4  & 88.7  & 66.0  & 66.8 \\
    SCSN~\cite{chen2020salience}  & 88.3  & 92.4  & 79.0  & 91.0  & 81.0  & 84.7 \\
    VT-ReID~\cite{xiang2024vt}  & 88.1  & 93.8  & 79.2  & \underline{92.6}  & 85.3  & 88.3 \\
    DMCL~\cite{xiang2024learning}   & \underline{92.1}  & \underline{96.3}  & \textbf{87.6}  & \textbf{93.5}  & \underline{86.5}  & \underline{89.7}  \\
    \midrule
    \textbf{GPF-Net}  & \textbf{93.1}  & \textbf{98.1}  & 74.4  & 74.1  & \textbf{88.2}  & \textbf{98.9}  \\
    \bottomrule
    \end{tabular}}%
  \label{tab3}%
\end{table}%

\begin{figure}[!t]
    \centering
    \includegraphics[width=0.5\textwidth]{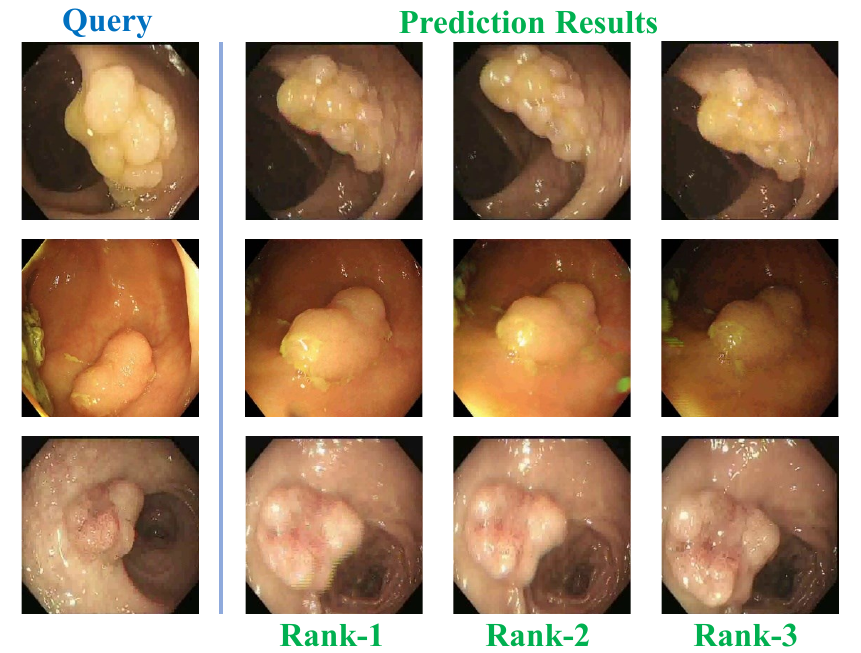}
    \caption{Examples of Top-3 ranking results for some query images of GPF-Net. In each row, images are arranged in descending order according to their similarities with the query on the left.}
    \label{fig3}
\end{figure}

\textbf{Person Re-Identification.}
To further prove the effectiveness and generalizablity of our method on other related object ReID tasks, we also compare our GPF-Net with existing methods in Table~\ref{tab3}. we can easily observe that our method can achieve the state-of-the-art performance on Market-1501, DukeMTMC-reID and CUHK03 datasets with considerable advantages respectively. For example, our GPF-Net method can achieve a mAP/Rank-1 performance of 93.1\% and 98.1\% respectively on Market-1501 dataset, leading \textbf{+1.8\%} improvement of Rank-1 accuracy when compared to the second best method DMCL~\cite{xiang2024learning}. In addition, our GPF-Net method can also obtain the improvement of \textbf{+9.2\%}  in terms of Rank-1 accuracy on CUHK03 dataset when compared to the DMCL~\cite{xiang2024learning}. However, we observe a notable shortcoming of GPF-Net on the DukeMTMC-reID dataset. We attribute this performance drop to the inherent class imbalance within the dataset, which predisposes the model to overfit frequent sample patterns while neglecting the distinctive features of under-represented instances, ultimately undermining both the fairness and representational completeness of the learned embeddings.

\begin{figure*}[!t]
    \centering
    \includegraphics[width=0.95\textwidth]{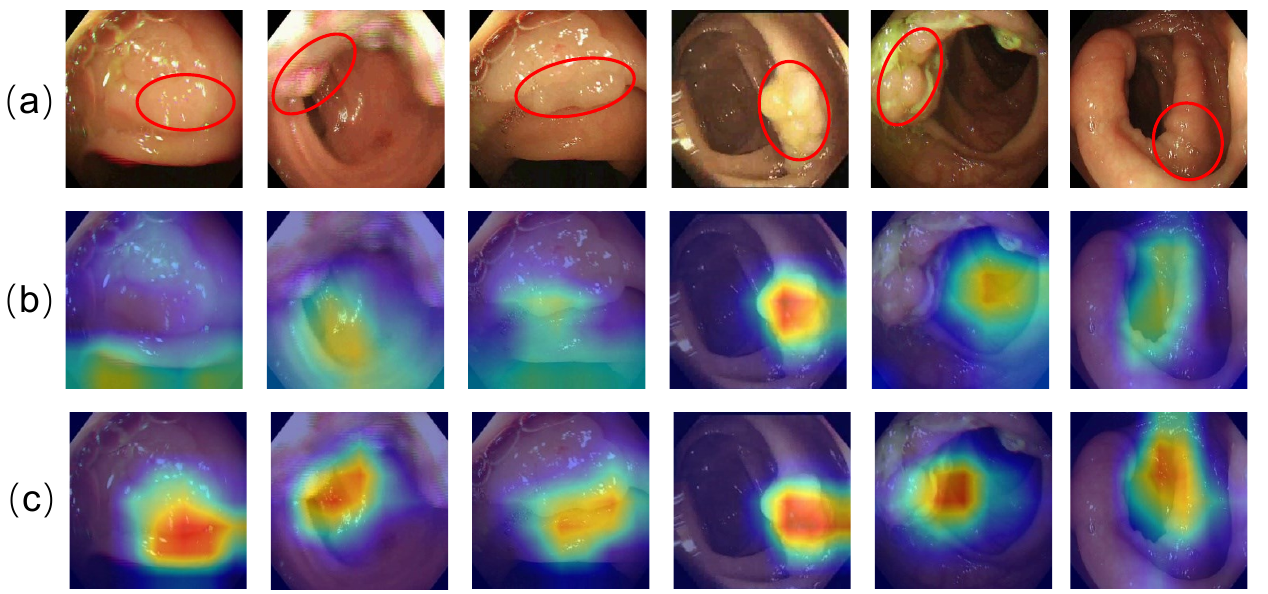}
    \caption{Visualization of attention maps of our GPF-Net under different input setting: (a) Original images; (b) CNN-based training method without text information; (c) Our gated progressive fusion strategy. Our model effectively learns global context and meaningful visual features, improving multimodal understanding and enabling automatic discovery of deeper, more representative image features.}
    \label{fig4}
\end{figure*}

\subsection{Ablation Studies}


\textbf{Effectiveness of  Multimodal Feature Fusion GPF-Net Framework:}
Firstly, from the quantitative aspect, we evaluate the effectiveness of multimodal feature fusion GPF-Net framework. As illustrated in Table~\ref{table1}, when adopting our gated progressive fusion framework, results show that mAP accuracy improves significantly from 27.94\% to 68.86\% on Colo-Pair dataset with visual-text induction. Additionally, similar improvements can also be easily observed in terms of Rank-1, Rank-5 evaluation metrics, leading to +60.64\% and +50.49\% respectively. These results prove that gated progressive fusion paradigm has a direct impact on downstream polyp ReID task.

Secondly, from the qualitative aspect,  we also give some qualitative results of our proposed gated progressive fusion  framework on multimodal polyp scenarios. For example,
Figure~\ref{fig3} provides some ranking results of GPF-Net, attention visualization of GPF-Net is also illustrated in Figure~\ref{fig4}. To be more specific, we can obviously observe that our model attends to relevant image regions or discriminative parts for making polyp predictions, indicating that GPF-Net can greatly help the model learn more global context information and meaningful visual features with better multimodal understanding, which provides the possibility to discover deeper layer of neural network and more representative features from images automatically.

\begin{table}[!t]
  \centering
  \caption{Ablation study of different pre-training settings from Colo-Pair dataset. Note that the pre-trained model is then fine-tuned on  dataset for downstream polyp ReID task.}
  \footnotesize
  \setlength{\tabcolsep}{1.65mm}{
    \begin{tabular}{lccccc}
    \toprule
    \multirow{2}[4]{*}{Pre-training} & \multirow{2}[4]{*}{Text data} & \multirow{2}[4]{*}{Image data} & \multicolumn{3}{c}{Colo-Pair$\uparrow$} \\
\cmidrule{4-6}          &       &       & mAP & Rank-1 & Rank-5    \\
    \midrule
    Baseline & $\times$     & $\times$     & 27.94  & 19.53  & 39.13   \\
    GPF-Net \textit{w/} text   & $\checkmark$     & $\times$     & 29.54  & 20.63  & 42.81   \\
    GPF-Net \textit{w/} image  & $\times$     & $\checkmark$     & 59.91  & 67.85  & 82.64    \\
    GPF-Net (Ours)  & $\checkmark$     & $\checkmark$     & 68.86  & 80.17  & 89.62  \\
    \bottomrule
    \end{tabular}}%
  \label{table1}%
\end{table}%

\begin{table}[!t]
  \centering
  \caption{Compared to other object retrieval methods in terms of Params(M) and FLOPs(G).}
  \footnotesize
  \setlength{\tabcolsep}{0.78mm}{
    \begin{tabular}{cccccc}
    \toprule
     Method     & ViT~\cite{caron2021emerging}   & CVRL~\cite{qian2021spatiotemporal}      & VT-ReID~\cite{xiang2024vt} & DMCL~\cite{xiang2024learning}  & GPF-Net (Ours) \\
    \midrule
    Params & 86.6  & 59.7    & 86.1  & 153.3 & \textbf{51.3} \\
    FLOPs  & \textbf{17.6}  & 115     & 50.9  & 68.7  & 71.5 \\
    \bottomrule
    \end{tabular}}%
  \label{table2}%
\end{table}%

\textbf{Effectiveness of Dynamic Gating Progressive Fusion Mechanism:}
In this section, we proceed to evaluate the effectiveness of dynamic multimodal training strategy by testing whether text modality or image modality matters. According to Table~\ref{table1}, our dynamic multimodal training strategy
GPF-Net with text representation (GPF-Net \textit{w/} text) can lead to a significant improvement in Rank-1 of +1.1\% on Colo-Pair dataset when compared with baseline setting.
Furthermore, when adopting image representation, our method (GPF-Net \textit{w/} image) can obtain a remarkable performance of 59.91\% in terms of mAP accuracy, leadig a significant improvement of \textbf{+30.37\%} when compared to GPF-Net \textit{w/} text.  In particular, we find that the missing low-level features can be fused into each feature level in the pyramid, which indicates that our module can well handle small and thin polyps in complex medical scenarios. To sum up,
the effectiveness of the dynamic multimodal training strategy can be largely attributed to that it enhances the discrimination capability of collaborative networks during multimodal representation learning, which is vital for polyp re-identification in general domain where the target supervision is not available.

\subsection{Parameter Analysis}
To go even further, we also elaborate a parameter analysis in Table~\ref{table2}, from which we can observe that GPF-Net can obtain a parameter value of 51.3M and 71.5G in terms of Params and FLOPs respectively, which is significantly lower than previous methods in terms of parameter quantity compared to existing mainstream methods, demonstrating excellent lightweight characteristics. Although the introduction of multimodal feature fusion has led to an increase in the model's computational complexity, especially in terms of FLOPs, the extent of performance improvement far outweighs the rise in computational cost.

\section{Conclusion}
 \label{sec5}
In this work, we present GPF-Net, a novel framework for colonoscopic polyp re-identification that addresses the limitations of single-stage multimodal fusion through a hierarchical gated progressive fusion strategy. By integrating a dynamic gating mechanism with multi-layer cross-modal interactions, our model effectively captures both fine-grained visual-text correlations and high-level semantic relationships. Importantly, our method ensures that complementary information from visual and textual modalities is progressively refined, from which we have proved that learning representation with multiple-modality can be competitive to methods based on unimodal representation learning.
In the future, we will incorporate additional modalities beyond visual and textual data, then integrate temporal information from consecutive video frames or depth maps from 3D colonoscopic reconstruction to further enhance polyp discrimination.

\bibliographystyle{IEEEbib}
\bibliography{refs}

\end{document}